\documentclass[conference]{IEEEtran}
\IEEEoverridecommandlockouts

\usepackage{hyperref}
\usepackage{cite}
\usepackage{amsmath,amssymb,amsfonts}
\usepackage{algorithmic}
\usepackage{graphicx}
\usepackage{textcomp}
\usepackage{xcolor}
\usepackage{multirow}
\usepackage{subfigure}
\usepackage{tikz}
\usepackage{fancyhdr}

\def\BibTeX{{\rm B\kern-.05em{\sc i\kern-.025em b}\kern-.08em
    T\kern-.1667em\lower.7ex\hbox{E}\kern-.125emX}}
\begin{document}

\title{Image-to-Joint Inverse Kinematic of a Supportive Continuum Arm Using Deep Learning
\thanks{}
}

\author{\IEEEauthorblockN{Shayan Sepahvand}
    \IEEEauthorblockA{Department of Mechanical,\\
    Industrial, and Mechatronics Engineering\\
    Toronto Metropolitan University\\
    Toronto, Canada\\
    Email: shayan.sepahvand@torontomu.ca}
    \and
    \IEEEauthorblockN{Guanghui Wang}
    \IEEEauthorblockA{Department of Computer Science\\
    Toronto Metropolitan University\\
    Toronto, Canada\\
    Email: wangcs@torontomu.ca}
    \and
    \IEEEauthorblockN{Farrokh Janabi-Sharifi}
    \IEEEauthorblockA{Department of Mechanical,\\
    Industrial, and Mechatronics Engineering\\
    Toronto Metropolitan University\\
    Toronto, Canada\\
    Email: fsharifi@torontomu.ca}
}

\makeatletter
\def\ps@IEEEtitlepagestyle{
    \def\@oddfoot{\customfootnote}%
    \def\@evenfoot{}%
}
\makeatother

\newcommand{\customfootnote}
{\footnotesize  \textbf{ \color{red}Accepted in 2024 21st Conference on Robots and Vision (CRV)} \href{https://doi.org/10.21428/d82e957c.d8706a7c}{https://doi.org/10.21428/d82e957c.d8706a7c}  \hfill}

\color{black}
\fancyhf{}
\renewcommand{\headrulewidth}{0pt}
\fancypagestyle{GlobalFootnote}{%
    \lfoot{\customfootnote} 
}
\pagestyle{GlobalFootnote} 


\maketitle

\begin{abstract}
In this work, a deep learning-based technique is used to study the image-to-joint inverse kinematics of a tendon-driven supportive continuum arm. An eye-off-hand configuration is considered by mounting a camera at a fixed pose with respect to the inertial frame attached at the arm base. This camera captures an image for each distinct joint variable at each sampling time to construct the training dataset. This dataset is then employed to adapt a feed-forward deep convolutional neural network, namely the modified VGG-16 model, to estimate the joint variable. One thousand images are recorded to train the deep network, and transfer learning and fine-tuning techniques are applied to the modified VGG-16
to further improve the training. Finally, training is also completed with a larger dataset of images that are affected by various types of noises, changes in illumination, and partial occlusion. The main contribution of this research is the development of an image-to-joint network that can estimate the joint variable given an image of the arm, even if the image is not captured in an ideal condition. The key benefits of this research are twofold: 1) image-to-joint mapping can offer a real-time alternative to computationally complex inverse kinematic mapping through analytical models; and 2) the proposed technique can provide robustness against noise, occlusion, and changes in illumination. The dataset is publicly available on \href{https://kaggle.com/datasets/5309cc77b7c5f20dee25301334d872f76fc91e53ac3c25521875b22957021f33}{Kaggle}.
\end{abstract}


\section{Introduction}
A continuum robot (CR) is characterized as a manipulable structure made of constituent material capable of assuming curved shapes with uninterrupted tangent vectors \cite{Burgner-Kahrs2015}. Taking inspiration from the flexible appendages of animals with flexible bodies, continuum robots employ a sequence of continuous curves as the structural basis for generating bending motion, as opposed to using rigid-link robots \cite{Jingyu2022}. Continuum robots (CRs) differ from hyper-redundant and pseudo-continuum robots due to the presence of continuous backbones, allowing their structure to be conceptualized as infinite degrees of freedom (DoF). This design offers several advantages, including dexterity, lightweight construction, simplicity \cite{Jones2006}, compliance, and scalability that allows small-scale construction to enable their operation in cluttered and constrained environments, which in turn leads to superior performance in interaction with humans and tasks where collaboration is required\cite{Janabi-Sharifi2021}. 

Leveraging their soft characteristics, they have gained widespread adoption in the field of medical surgery \cite{Yang2023}, industry operation \cite{wu2019unsupervised}, subsea applications \cite{Davies1998}, repair of gas turbine engines in on-wing scenarios\cite{Dong2017}, and search and rescue operations \cite{Tsukagoshi2001}. Despite these advancements, CRs are required to be further developed to improve their maneuverability, payload, and speeds for potentially complex tasks such as manipulation and grasping. Consequently, more accurate modeling and real-time control systems are required to deal with these problems. Efficient kinetic-dynamic mapping techniques are hence at the center of attention of research on CRs. Among those, kinematic mapping takes priority due to the quasi-static nature of motions in CRs. Particularly, the inverse kinematic (IK) solution is often needed for position control of CRs for a given end-point trajectory.  The IK  problem comprises obtaining the joint variables corresponding to a given end-effector pose \cite{Siciliano2009}. There are two general approaches to solving the IK problem. The commonly used IK method based on differential kinematics requires the calculation of the analytical Jacobian to solve the problem. However, analytical Jacobian is not readily available in exact CR models (using Coserrat rod theory), which require the solution of partial differential equations (PDEs), not leading to an efficient and closed-form solution. A vision-based learning approach seems to offer an alternative approach.


Recently, learning-based methods have been proposed to deal with the problem of robots' forward and inverse kinematics. For instance, in \cite{Elkholy2020},  the IK of a manipulator with seven DoF employing Artificial Neural Networks (ANNs) is solved. The CNN takes the end-effector's desired pose as the input and generates the joint variable of each joint. Grassmann et al. suggested an approach to solving the FK and the IK of concentric tube CRs \cite{Grassmann2018}. Here, two ANNs with Rectified Linear Unit (ReLU) activation functions are suggested to estimate the kinematics \cite{Grassmann2018}. The problem is further formulated for reinforcement learning (RL) to provide a model-free RL-based approach for IK of concentric tube CRs \cite{Iyengar2020}. The most analogous work to ours is presented in \cite{Liang2021}, where a methodology is used to obtain the IK for concentric tube CRs. They showed that the learning method outperforms the vanilla numerical technique in terms of accuracy.

All of the above methods addressed the learning-based modeling of the IK for rigid-link arms or concentric tube CRs. However, the problem of learning-based IK  for the supportive tendon-driven cooperative CRs (SCCRs) \cite{Lotfavar2017} has not been investigated so far to the best of the author's knowledge. The existing gap motivated us to develop a vision-based learning technique for the IK of SCCRs.  The approach exploits the capability of deep CNNs, transfer learning, and fine-tuning to detect the features more efficiently and precisely estimate the joint variables. This is the main contribution of this paper. 

The content of this paper is as follows. In section \ref{s2}, the continuum arm's dynamic model is introduced briefly to justify the use of the learning-based method. The prototype specifications are explained in detail. The process of creating the dataset, the CNN structure, and its training are discussed in section \ref{s2}. Section \ref{s3} explains the experiments and the analysis in general. The learning curves are investigated, and the result of the prediction for the test dataset is also examined. We also studied the robustness of the method against noise and illumination variations and used visual inspection of features to analyze the cons and pros of the convolutions. Finally, the concluding remarks are presented in section \ref{s4}. 

\section{Methodology} \label{s2}
\begin{figure}
    \centering
    \includegraphics[width = 0.5\textwidth]{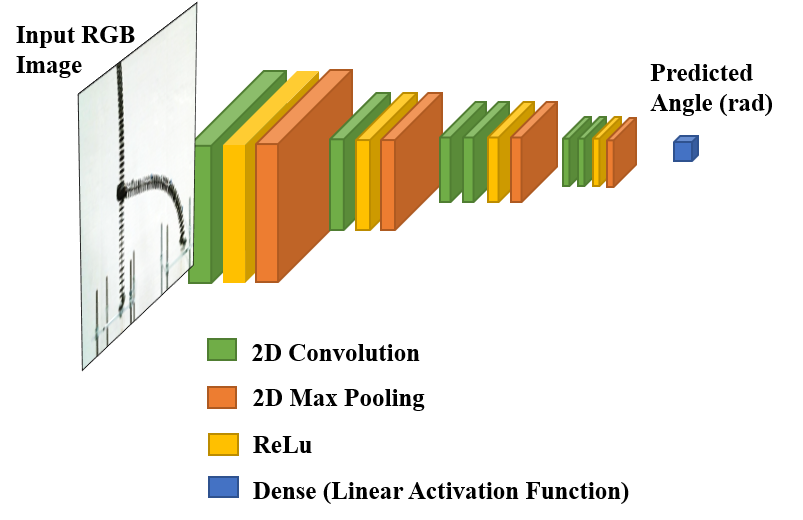}
    \caption{Architecture of modified VGG-16 model}
    \label{fig:Model Structure}
\end{figure}

In this section, a methodology is suggested to approximate the IK for a supportive tendon-driven continuum robot from an image using a deep learning-based approach. A deep CNN is employed to learn the joint variables using the visual information. A simple webcam is used to acquire the frames by a camera to construct the training dataset. Experimental tests are used to verify the performance of the proposed method by evaluating the precision of the estimations and robustness against factors such as noise and change in illumination. The model is shown in Fig. \ref{fig:Model Structure}.

\subsection{Dynamics of the supportive tendon-driven continuum robot}
In this paper, a supportive tendon-driven CCR is considered. The following system \eqref{eq:eq1} of nonlinear PDEs describe the general dynamics of the system using the Cosserat rod modeling approach \cite{Janabi-Sharifi2021}: 
  
\begin{equation}
\begin{split}
    \frac{\partial \xi}{\partial s} &= f\left(\frac{\partial \xi}{\partial t}, \xi, \tau, t\right)\\  
    \frac{\partial \xi}{\partial t} &= g\left(\frac{\partial \xi}{\partial t}, \xi, \tau, t\right)
\end{split}
\label{eq:eq1}
\end{equation}

\noindent where $\tau$ is the vector of the tendon's tension usually applied at the tip, changing the shape of the CR, $\xi$ is the vector of states, $s$ denotes the spatial variable, $t$ is the time, and $f(.)$ and $g(.)$ are two nonlinear, complex functions. The details of Cosserat rod modeling are provided in the literature \cite{Janabi-Sharifi2021}. It is worth highlighting that the exerted tension is applied through DC motors. Even though the displacement command of the DC motors does not appear in \eqref{eq:eq1}, a simple look-up table can be used to map the DC motor variables to the applied tension at the tip. This also omits the need for precise and expensive force sensors to measure the tension.  

\subsection{Development and Design of the Prototype}

\begin{figure*}
    \centering
    \includegraphics[width = \textwidth]{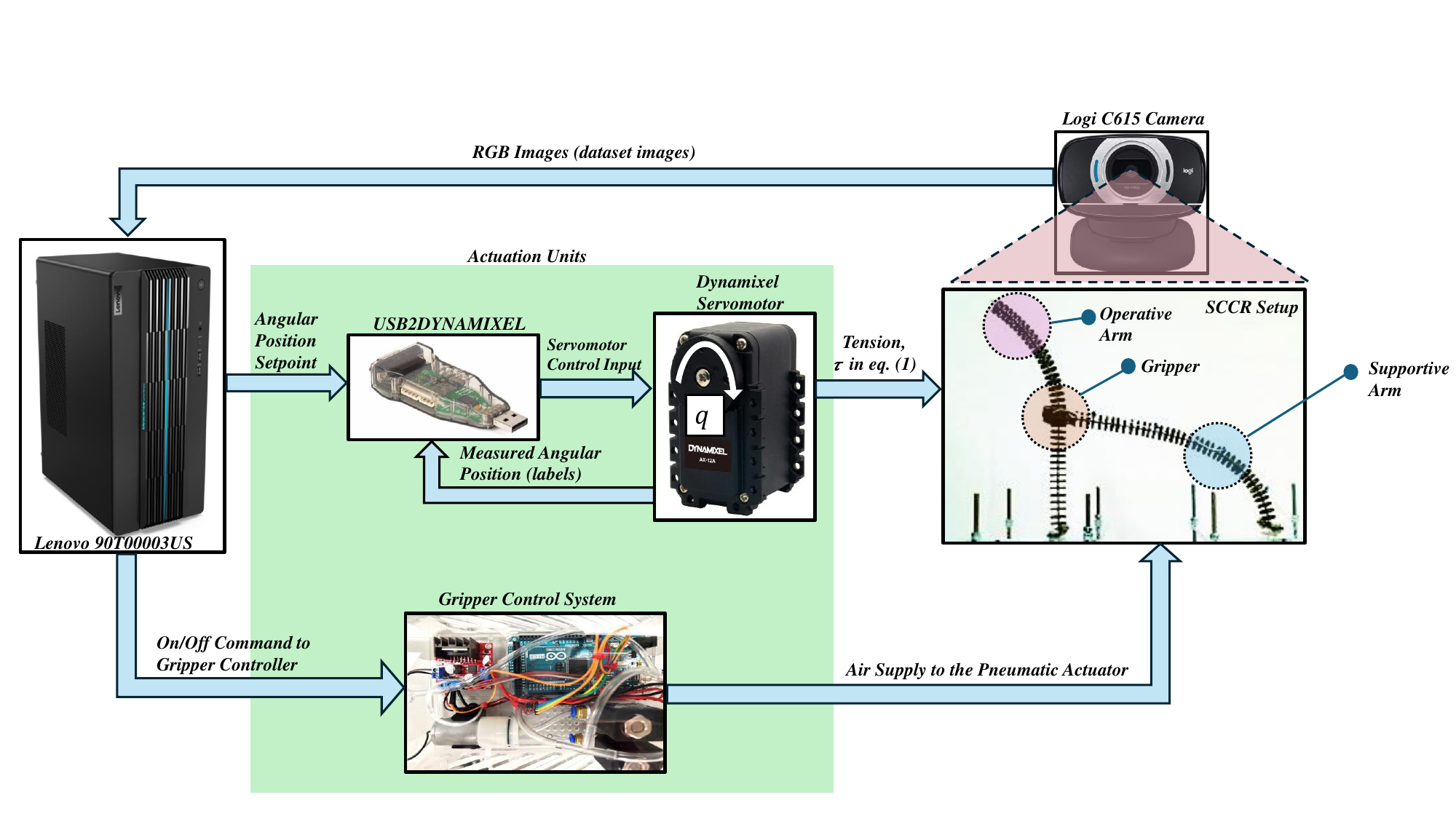}
    \caption{The experimental setup used to create the dataset}
    \label{fig:SCR Setup}
\end{figure*}

As illustrated in Fig. \ref{fig:SCR Setup}, the SCCR comprises the operative (the arm on the left) and supportive arms (the arm on the right). Each arm consists of several elements: additional spacer disks for transmitting motor tensions to the tip flexible spring steel backbone, spacer disks made with PLA filament to guide the tendons, and four braided Kevlar lines serving as tendons. These tendons align along a circle with a radius of $18.55$ (mm), running parallel to the backbone. The manipulation of the tendons is done using Dynamixel servomotors AX-12A (Robotis, Lakeforest CA). A pneumatic gripper is mounted on the top of the supportive arm to grip the operative arm. The specification of the prototype is summarized in Table \ref{tab:PS}.

\begin{table}[]
    \centering
        \caption{The experimental setup and its components' specifications.}
    \begin{tabular}{c c c}
         \textbf{Component} & \textbf{Specification} & \textbf{Value}\\ 
         \hline
         \multirow{7}{*}{Camera}
         &Image Size &$1080 \times 720$ (pixels)\\
         &Spatial Resolution & 96 (dpi)\\
         &Intensity Resolution & 256\\
         &Channels&RGB\\
         &Autofocus&Off\\
         &Frame Rate&$30$ (fps)\\
         &Field of View& $78^\circ$\\
         \hline
         Tendon&Breaking strength& 31.75 (Kg) \\
         \hline 
         \multirow{4}{*}{Backbone}
         &Density& 7800 $\mathrm{(Kg/m^3)}$\\
         &Young’s modulus (E) & 207  $\mathrm{(GPa)}$\\
         &Length (L) & 0.4 $\mathrm{(m)}$\\
         &Radius (r) & 0.9 $\mathrm{(mm)}$\\
         \hline
    \end{tabular}
    \label{tab:PS}
\end{table}

\subsection{Data Acquisition}
\begin{figure}
  \centering

  \subfigure[$q = 2.063 \mathrm{(rad)}$]{\includegraphics[width=0.3\linewidth]{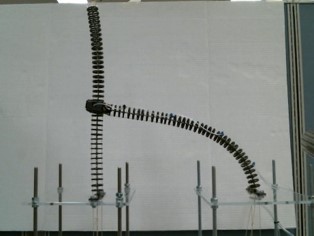}}
  \subfigure[$q = 2.902 \mathrm{(rad)}$]{\includegraphics[width=0.3\linewidth]{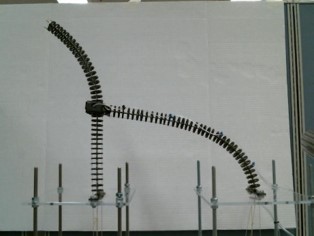}}
  \subfigure[$q = 1.688 \mathrm{(rad)}$]{\includegraphics[width=0.3\linewidth]{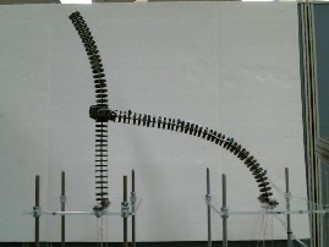}}
  
  \subfigure[$q = -2.386\mathrm{(rad)}$]{\includegraphics[width=0.3\linewidth]{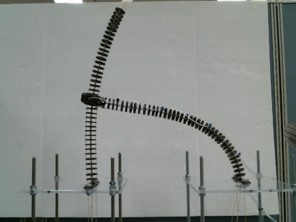}}
  \subfigure[$q = 2.675 \mathrm{(rad)}$]{\includegraphics[width=0.3\linewidth]{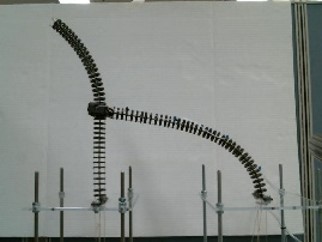}}
  \subfigure[$q = 0.265 \mathrm{(rad)}$]{\includegraphics[width=0.3\linewidth]{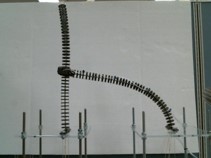}}

  \caption{Six images of the training dataset and their labels}
  \label{fig:dataset_images}
\end{figure}

Data acquisition involves several steps. Firstly, an ON/OFF command is sent to the Arduino; accordingly, the pneumatic actuators direct the compressed air to the pneumatic gripper so the supportive arm grips the operative arm. Then, a random joint variable command is transmitted to the Dynamixel motors using MATLAB Software Development Kit (SDK). This joint variable serves as the label used to train the CNN, which is subsequently stored in a text file once it is sent to the servo motor After that, the Logitech C615 HD 1080p camera captures an image and stores it in the dataset directory; this process is repeated until the pre-specified number of images is acquired. However, there is a delay (about two seconds) between each iteration to allow the SCR to reach its final configuration as well as the PC to communicate with the control boards and camera. Fig. \ref{fig:dataset_images}, indicates six training images and their corresponding joint variables $q$ shown in Fig. \ref{fig:SCR Setup}.

\subsection{Image-to-joint CNN Training}
The input of the CNN is an RGB image, and it outputs the angle of the actuator. The model used is a modified VGG-16, and the base model is trained using the ImageNet dataset \cite{Deng2009}. The last block containing two 2D convolutional, a ReLu, and a 2D max pooling layer, is discarded from the conventional VGG-16 to preserve more features since the pre-trained kernels do not suit our application due to the fact that these filters indicate their best performance for the ImageNet dataset, not ours. Next, using transfer learning techniques, the convolutional layer weights are frozen, and the prediction layers are omitted and replaced with a dense layer with a linear activation function. However, to further enhance the results, after adjusting the weights of the prediction layer, its weights can be frozen. In contrast, the weights of the convolutional layers are unfrozen and trained with a lower learning rate for fewer epochs compared to the initial training, as the performance of the CNN is drastically impacted by changes in the weights of the kernels. The last layer is a single neuron with a linear activation function, which outputs the joint angle $q$. The model structure is summarized in Table \ref{tab:model structure}.

\begin{table}[]
    \centering
    \caption{The structure of the model for a $600\times600$ RGB input image}
    \begin{tabular}{c c c c}
    \hline
        \textbf{Layer} & \textbf{Type} & \textbf{Input Shape} & \textbf{Output Shape} \\
        \hline\hline
        1& Input &$(\textsc{BS}^*,300,300,3)$ & $(\textsc{BS},300,300,3)$\\
        \hline
        2& 2D Convolution &$(\textsc{BS},300,300,3)$ & $(\textsc{BS},300,300,64)$\\
        \hline
        3& ReLu &$(\textsc{BS},300,300,64)$ & $(\textsc{BS},300,300,64)$\\
        \hline
        4& 2D Max Pooling &$(\textsc{BS},300,300,64)$ & $(\textsc{BS},150,150,64)$\\
        \hline
        5& 2D Convolution &$(\textsc{BS},150,150,64)$ & $(\textsc{BS},150,150,128)$\\
        \hline
        6& ReLu &$(\textsc{BS},150,150,128)$ & $(\textsc{BS},150,150,128)$\\  
        \hline      
        7& 2D Max Pooling &$(\textsc{BS},150,150,128)$ & $(\textsc{BS},75,75,128)$\\  
        \hline      
        8& 2D Convolution &$(\textsc{BS},75,75,128)$ & $(\textsc{BS},75,75,256)$\\
        \hline
        9& 2D Convolution &$(\textsc{BS},75,75,256)$ & $(\textsc{BS},75,75,256)$\\  
        \hline   
        10& ReLu &$(\textsc{BS},75,75,256)$ & $(\textsc{BS},75,75,256)$\\  
        \hline 
        11& 2D Max Pooling &$(\textsc{BS},75,75,256)$ & $(\textsc{BS},37,37,256)$\\  
        \hline         
        12& 2D Convolution &$(\textsc{BS},37,37,256)$ & $(\textsc{BS},37,37,512)$\\  
        \hline  
        13& 2D Convolution &$(\textsc{BS},37,37,5)$ & $(\textsc{BS},37,37,512)$\\  
        \hline 
        14& ReLu &$(\textsc{BS},37,37,512)$ & $(\textsc{BS},37,37,512)$\\  
        \hline 
        15& 2D Max Pooling &$(\textsc{BS},37,37,512)$ & $(\textsc{BS},18,18,512)$\\  
        \hline 
        16& Flatten &$(\textsc{BS},18,18,512)$ & $(\textsc{BS},165888)$\\
        \hline 
        17& Dense (Linear) &$(\textsc{BS},165888)$ & $(\textsc{BS},1)$\\               
        \hline\hline
        \multicolumn{4}{l}{*BS denotes the batch size.}
    \end{tabular}
    \label{tab:model structure}
\end{table}

\section{Results} \label{s3}

The model implemented in TensorFlow 2.13.1 and is trained on a dataset $\{\mathbb{I},\mathbb{Q}\}$ where $\mathbb{I}$ denotes the collection of 1000 images of the SCCR setup, and $q_i \in \mathbb{Q}$ with \( q_i \in \left[\frac{-7\pi}{2}, \frac{7\pi}{2}\right] \)  represents a scalar angle sample. The data split ratio for learning, validation, and testing is 8:1:1. During transfer learning and fine-tuning stages, the loss function, which is the mean squared error (MSE), is minimized. The performance of the system against common disturbing factors such as different types of noise, partial occlusion, and change in illumination is studied. A statistical analysis of the results of each stage is conducted to give a general view of the performance of the system. Eventually, to validate the results further, a K-fold cross-validation technique is established.

\subsection{Learning Curves}
\begin{figure}
  \centering

  \subfigure[The transfer learning stage]{\includegraphics[width=0.45\linewidth]{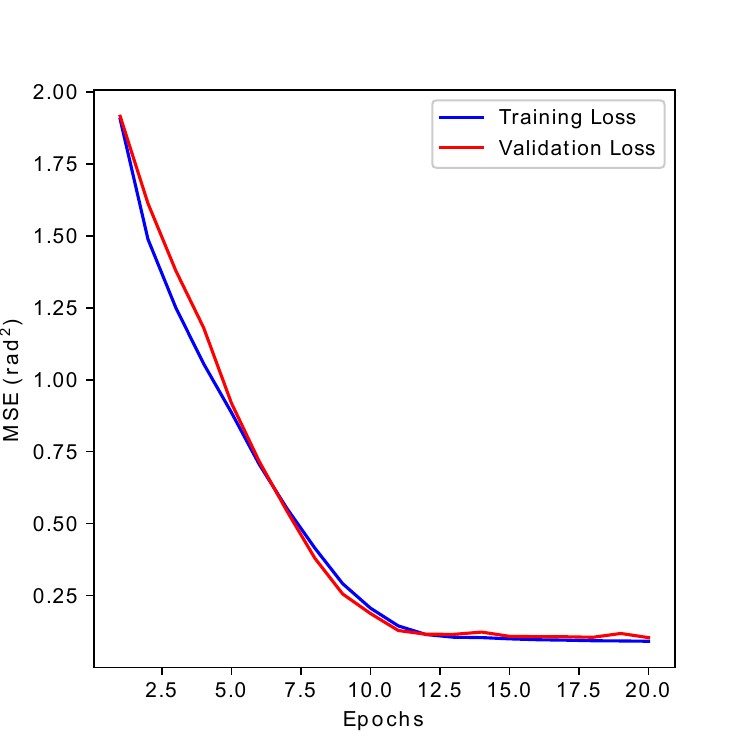}} 
  \subfigure[The fine-tuning stage]{\includegraphics[width=0.45\linewidth]{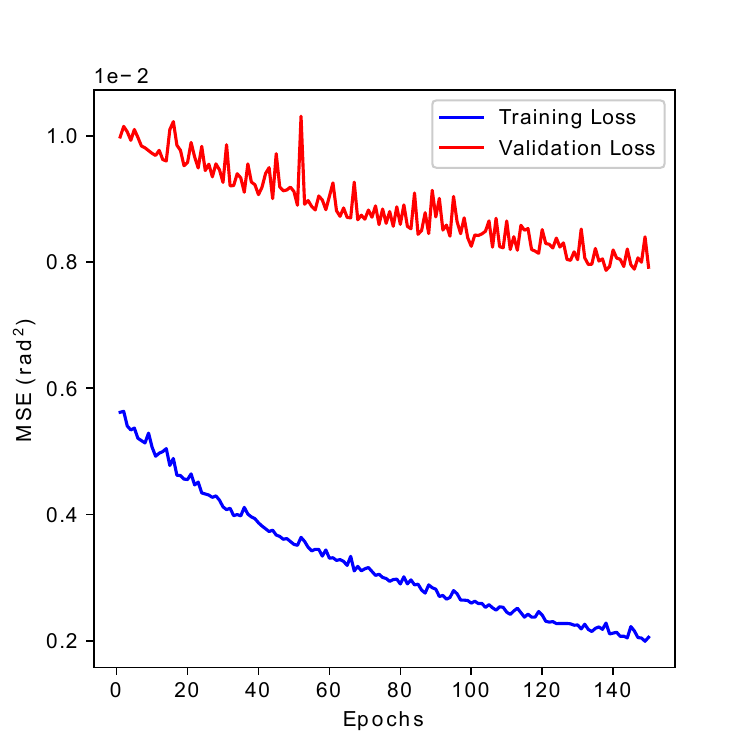}}
  
  \caption{The learning curves }
  \label{fig:LearningCurves}
\end{figure}

In this section, the learning curves of the modified VGG-16 for $800$ training images are presented. Nonetheless, before using these images for training, they were resized to the resolution of $300 \times 300$ pixels using nearest neighborhood interpolation to avoid memory limitations. In addition, as the intensity resolution is $256$, the image is divided by $255$ for normalization. The Adam optimizer with learning rate $\lambda = 10^{-6}$, decay rates $0.9$, and $0.999$, a mini-batch size of $16$, and the epoch number $150$ are selected as the hyper-parameters. Figs. \ref{fig:LearningCurves} (a) and (b) show how training and validation losses change with epochs for the two stages of training. During the first step, the training and validation losses undergo a similar descending trend, dropping from approximately $2$ to less than 0.25 $\mathrm{rad}^2$ in 20 epochs. Likewise, the pattern is descending for the second stage; nonetheless, the change in loss is infinitesimal. The training loss decreased from 0.005 to 0.002 $\mathrm{rad}^2$; nonetheless, the validation loss exhibits fewer variations and lowered from $0.01$ to approximately 0.008 $\mathrm{rad}^2$.

\subsection{Prediction results for the test images}
\begin{figure}
    \centering
\includegraphics[width = 0.45\columnwidth]{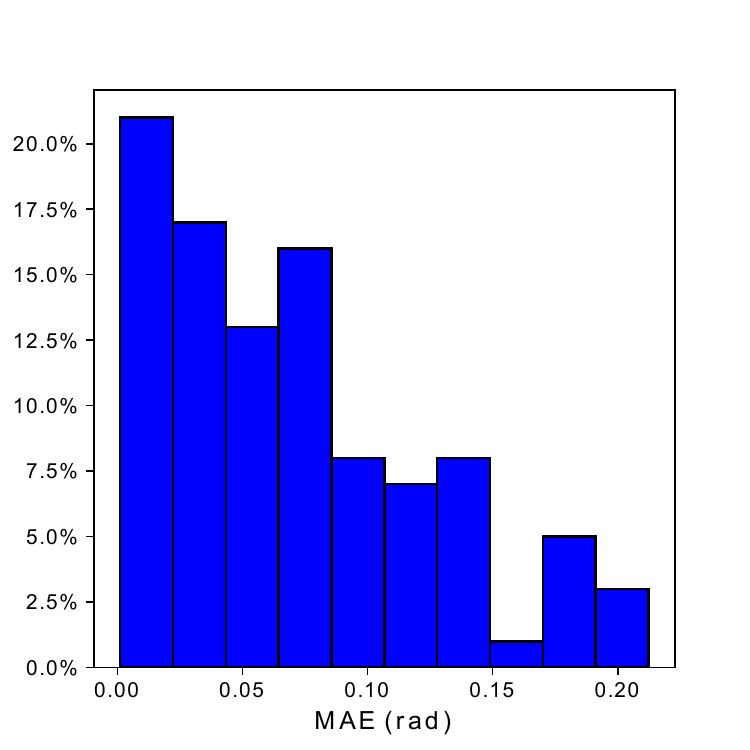}
  \caption{Prediction error distribution for the image-to-joint CNN}
  \label{fig:ErrorDistribution}

\end{figure}

The error distributions for $100$ unforeseen test images using the CNN  are illustrated in Fig. \ref{fig:ErrorDistribution}. The histogram is left-skewed, suggesting that most of the error values lie on the lower end, but there are a few that are larger. Just above $40\%$ of the errors are less than $0.05$ rad, whereas about $10\%$ are larger than $0.15$ rad. In terms of the estimation time for the test images, the average time was \(0.7\) s when computation was performed on the GeForce RTX 3060.

\subsection{Comparison between different CNNs}

In this section, the test results are investigated for various state-of-the-art CNN models in terms of the loss value of test data after undergoing $150$ epochs of learning. The results of this experiment, along with the number of layers of each CNN, are listed in Table \ref{tab:diff_cnns}. The best estimation error is provided by DenseNet201 with a loss of $0.0106$, whereas EfficientNetB0 indicates poor estimation results. The loss values for DenseNet121 and InceptionV3 stand at $0.0198$ and $0.0286$, respectively. Overall, when the number of layers is considered as a deciding factor affecting the computation time, VGG16 outperforms the other CNNs in our experiment in terms of the loss value and simplicity. 

\begin{table}[]
    \centering
    \caption{A comparison between different CNNs}
    \begin{tabular}{c c c c}
        \textbf{CNN} & \textbf{MSE} & \textbf{MAE} & \textbf{Layers} \\
        \hline
        InceptionV3& $0.0286$ &$0.1332$ & $189$\\
        \hline
        DenseNet201& $0.0106$ &$0.0846$ & $402$\\
        \hline 
        DenseNet121& $0.0198$ &$0.121$ & $242$\\
        \hline         
        EfficientNetB0& $4.1208$ &$1.7535$ & $132$\\
        \hline         
        MobileNet& $0.0592$ &$0.1942$ & $55$\\        
    \end{tabular}
    \label{tab:diff_cnns}
\end{table}

\subsection{Robustness tests on training images}
\begin{figure}
  \centering

  \subfigure[Gamma transformed image with $\gamma = 0.5$]{\includegraphics[width=0.4\linewidth]{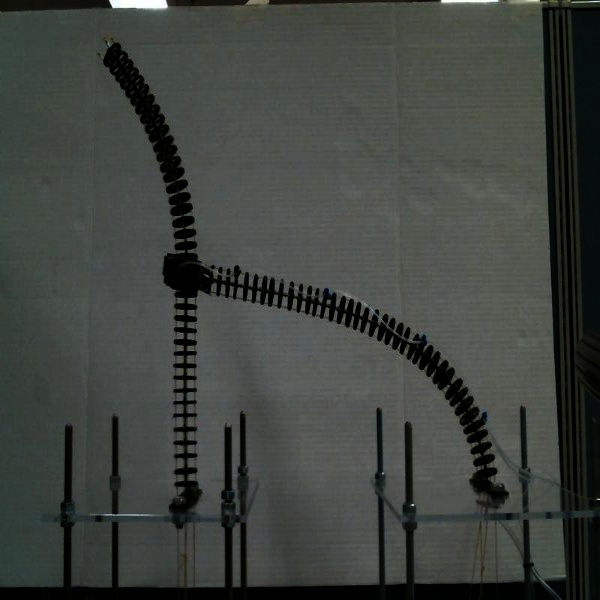}} 
  \subfigure[Gamma transformed image with $\gamma = 2$]{\includegraphics[width=0.4\linewidth]{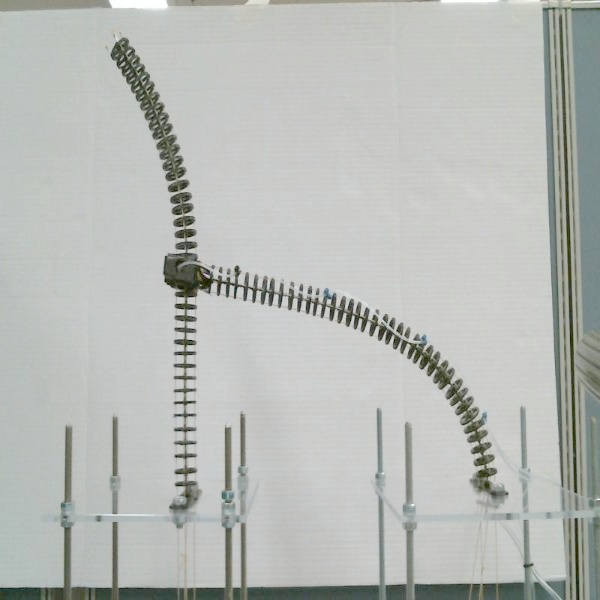}}

  \caption{Sample of training images affected by change in illumination}
  \label{fig:syn_imgs}
\end{figure}

In real-world applications, the input images captured by the camera are affected by various disturbing factors such as noise and variation in illumination. In this section, the CNN is first trained with $7000$ images (in addition to the original images) that were affected by undesirable factors, including Gaussian with zero mean and standard deviation $0.01$, speckle noise with the same mean and standard deviation as the Gaussian noise, impulse noises with probability $0.01$ when its distribution is normal, low and high illumination, and partial occlusion. The gamma transformation with $\gamma = 0.2$ and $\gamma = 2$ was used to create dark and bright images to study the effect of illumination, as shown in Figs. \ref{fig:syn_imgs} (a) and (b), respectively. The images were partially occluded by squares with sides of $200$ pixels located at random coordinates in the image. The trained CNN was tested six times against 100 unforeseen images. Each test was designed to evaluate the performance of the trained CNN for one of the aforementioned factors. The same test images were employed to test the prediction of the CNN-trained original compared to the one that was adapted using the augmented dataset. The evaluation loss of the CNN trained with augmented images was lower than that of the original CNN, indicating the improvement in robustness against all of the factors. For example, in the case of the occurrence of Gaussian noise, the MAE significantly decreased from above $1.4$ to just above $0.2$ rad. The smallest change was for the occluded images, where the MAE dropped from $1$ to $0.4$ rad. 

\begin{figure}[t]
    \includegraphics[width=\columnwidth]{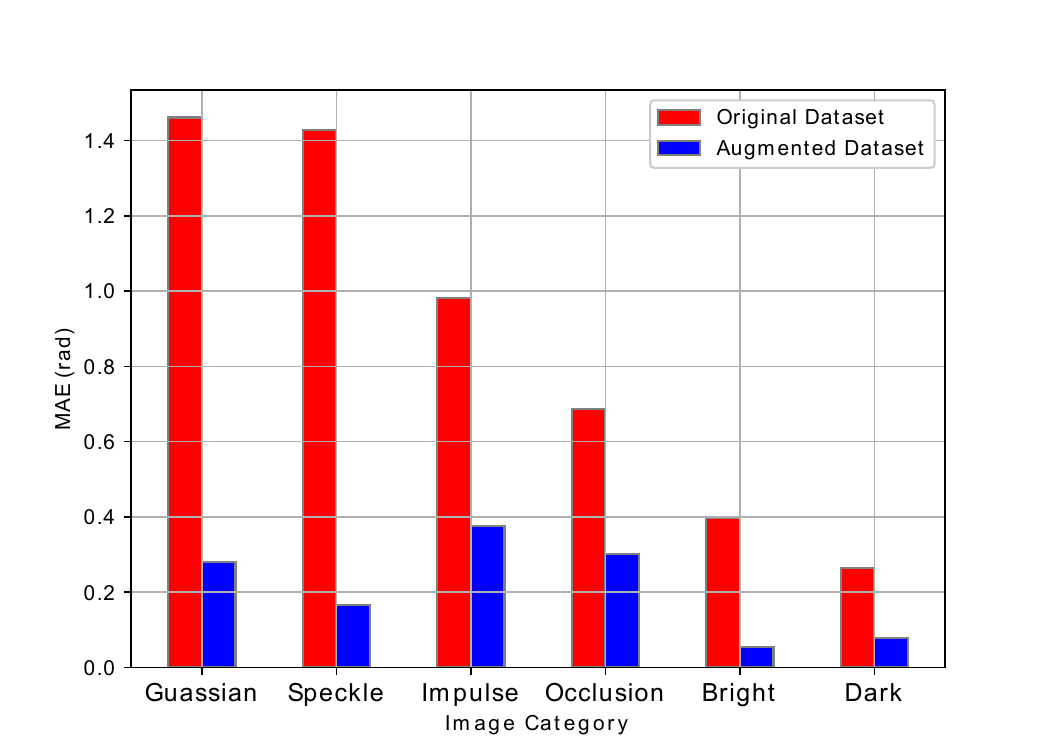}
    \caption{The bar plot of estimation MAE for different tests}
    \label{fig:violin plot}
\end{figure}

\subsection{K-fold cross-validation}

\begin{figure}
    \centering
    \includegraphics[width=1\columnwidth]{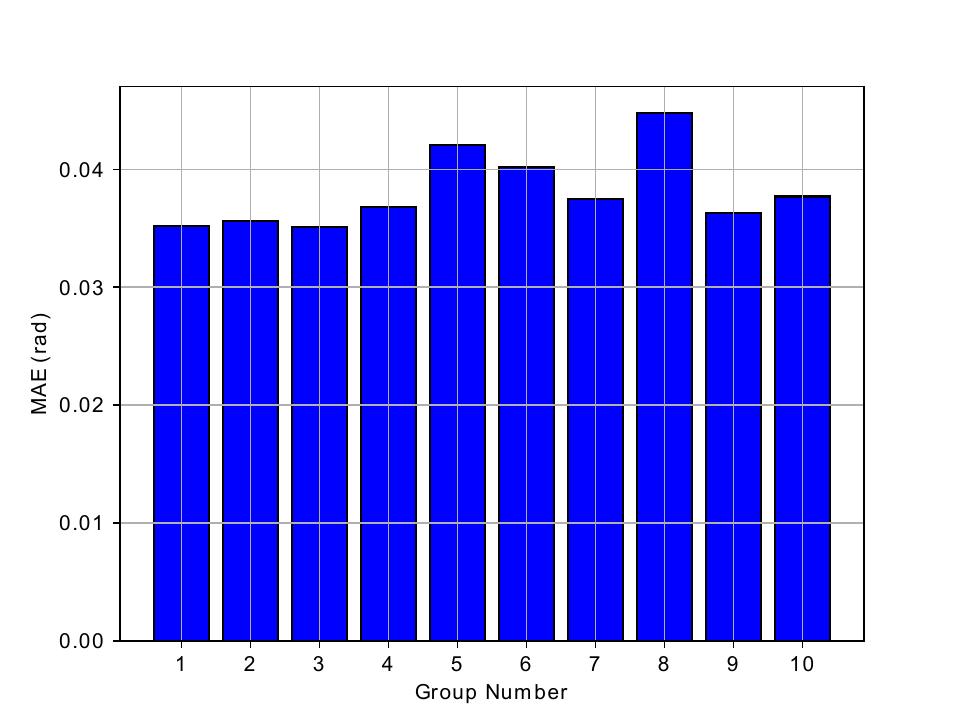}
    \caption{The bar plot of the MAE for all of the groups}
    \label{fig:cross-validation-violin}
\end{figure}

To further evaluate the performance of the CNN, the dataset is divided into ten groups of $100$ images. As a result, the model is trained ten times using nine subsets and tested against the remaining images. It is worth highlighting that as the images captured exhibit the random movement of the arm, the shuffling of the images is skipped. The distribution of the MAE is illustrated in Fig. \ref{fig:cross-validation-violin}. The distribution of the error in all groups is similar, suggesting the lack of over-fitting. Most of the errors lie between $0.03$ to $0.04$ rad, and the variance of the MAE for all groups is $9.440 \times 10^{-6}$ rad$^2$, confirming the consistency of the results for all of the groups.

\subsection{Visual inspection of feature maps}

\begin{figure}[th]
    \centering
    \includegraphics[width = 0.45\textwidth]{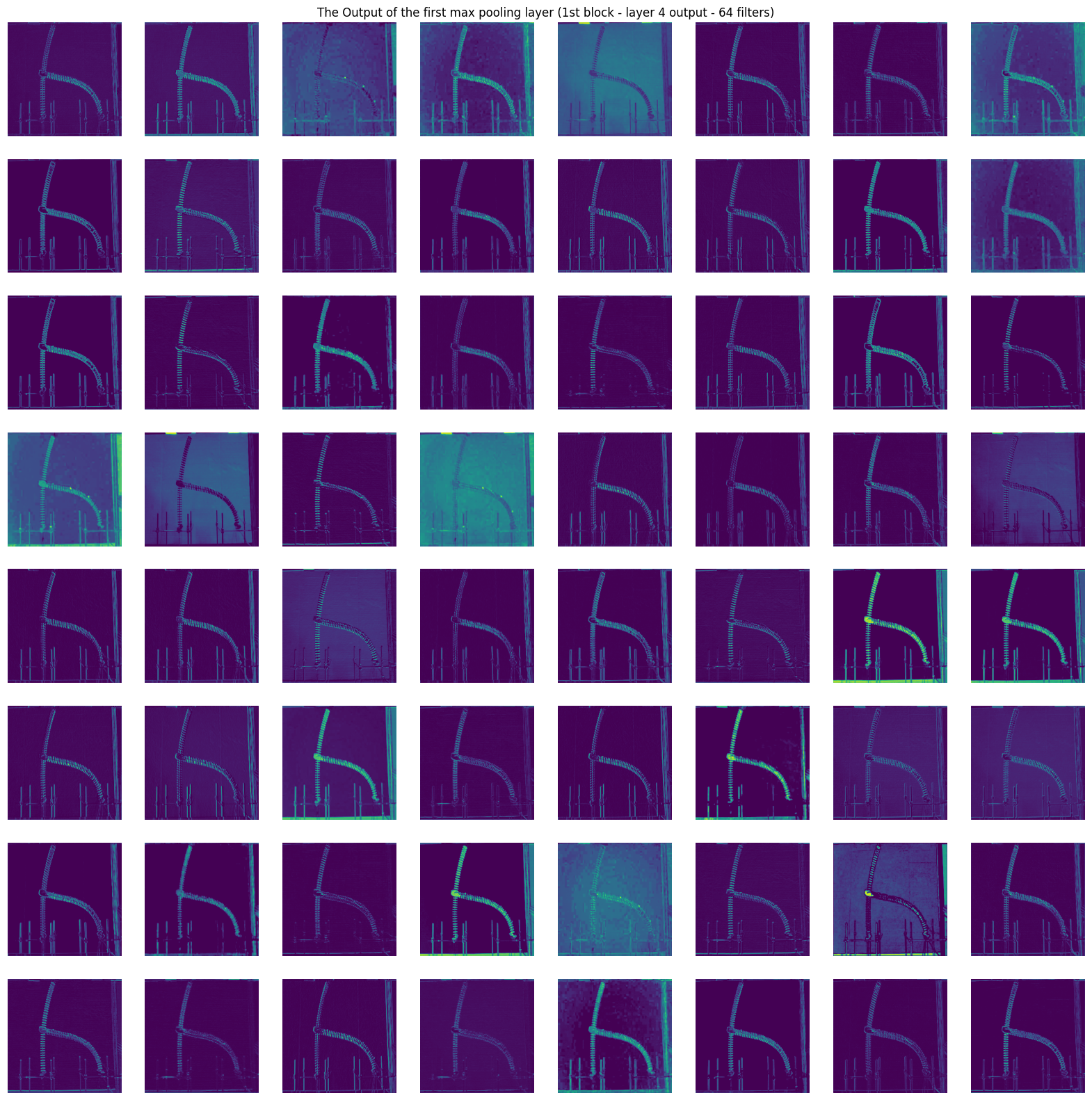}
    \caption{The output of the first block of the CNN (4th layer) }
    \label{fig:featuremap}
\end{figure}

One of the methods to analyze the performance of the deep CNN is to extract the feature maps that each block of the modified VGG-16 outputs. To this end, the output of the first block is shown in Fig. \ref{fig:featuremap}. Some of the kernels have a satisfactory performance in extracting the overall shape of the continuum arm. Nevertheless, some of these filters have a negative impact on the angle estimation. For example, several filters are trained to detect the background instead of the robot. Likewise, some filters detect the lines, like the line on the right side of the scene. The base of the robot is also detected in several convolutions, which is not desirable. Interestingly, there are filters that can be used to detect the point that supportive arm and operative arm connect.

\section{Conclusion}\label{s4}
In this paper, the problem of IK modeling of a supportive cooperative continuum robot is investigated. The image dataset is created using a USB webcam fixed in the room. Then, for each random joint variable, we captured an image and stored it along with the joint angle. After creating the dataset, we examined different techniques for choosing the appropriate deep CNN. As a result, a modified VGG-16 trained with the ImageNet dataset was selected. The last block and prediction layer are omitted since they did not serve as beneficial for our application. After the transfer learning stage, we utilized fine-tuning to further reduce the loss function final value. The distribution of the estimation error shows that about $40\%$ of the errors are less than $0.05$ rad. The robustness tests showed that the CNN is highly robust against different types of noises, changes in illumination, and partial occlusion as the mean absolute value of prediction error was less than $0.4$ rad for all cases. In addition, the K-fold cross-validation verified the lack of over-fitting for all groups. Lastly, we analyzed one of the feature maps to see whether all of the filters are useful, and we noticed that some of them do not contribute to the detection of the CR body, which is crucial for the estimation of IK.

\bibliographystyle{ieeetr}
\bibliography{Draft}

\end{document}